\documentclass[letterpaper, 10 pt, conference]{article}   
\usepackage{graphicx}
\usepackage{subcaption}
\usepackage{amsmath}
\usepackage{amssymb}
\usepackage{float}
\usepackage{url}
\graphicspath{ {images/} }

\newcommand{\nostarnote}[1]{}

\title{\LARGE \bf
SafeDrive: A Robust Lane Tracking System for Autonomous and Assisted Driving Under Limited Visibility
}

\author{Junaed Sattar \& Jiawei Mo\\
  Department of Computer Science, University of Minnesota,
  \\200 Union St SE, Minneapolis, MN, 55455, USA\\
{\tt\small \{junaed, moxxx066\} at umn.edu.}}

\begin{document}

\maketitle
\thispagestyle{empty}
\pagestyle{empty}

\begin{abstract}

We present an approach towards robust lane tracking for assisted and autonomous driving, particularly under poor visibility. Autonomous detection of lane markers improves road safety, and purely visual tracking is desirable for  widespread vehicle compatibility and reducing sensor intrusion, cost, and energy consumption. However, visual approaches are often ineffective because of a number of factors, including but not limited to occlusion, poor weather conditions, and paint wear-off. Our method, named SafeDrive, attempts to improve visual lane detection approaches in drastically degraded visual conditions without relying on additional active sensors. In scenarios where visual lane detection algorithms are unable to detect lane markers, the proposed approach uses location information of the vehicle to locate and access alternate imagery of the road and attempts detection on this secondary image. Subsequently, by using a combination of feature-based and pixel-based alignment, an estimated location of the lane marker is found in the current scene. We demonstrate the effectiveness of our system on actual driving data from locations in the United States with Google Street View as the source of alternate imagery. 

\end{abstract}

\section{Introduction}
\label{sec:introduction}
Recent advances in affordable sensing and computing technologies have given new impetus towards commercialization of a wide variety of intelligent technologies. A major consumer-targeted application has focused on increasing autonomy in transportation systems, the most prominent of which is the area of self-driven cars. Autonomous driving has been a key focus in both academic and industrial research and development activities~\cite{Thrun2010TRC} of late. Alongside fully autonomous commercial vehicles, mainstream auto manufacturers are equipping their vehicles with more intelligent technology with semi-autonomous, \emph{assistive} features -- the primary focus being increased safety. Many recent consumer-grade vehicles come with a number of such safety-enhancing features -- \emph{e.g.}, lane assist, blind-spot detection, radar-assisted braking, visual collision avoidance, driver fatigue detection~\cite{wang2006driver} -- with the number and quality of features increasing in higher-end, more expensive vehicles. These features are also available as add-on options, often costing a few thousand US dollars to install in a vehicle. Even then, not all vehicles are capable of fitting such a system, as these options require specific vehicle data and power interfaces, limiting their application to newer vehicles. However, to minimize distracted driving (which has approximately 20 {\em per cent} contribution to fatalities on the road~\cite{distracteddata2012}) and improve safety, many consumers are opting to buy newer vehicles with these features pre-installed. 


\begin{figure}[t!]
    \vspace{2mm}
    \centering
    \begin{subfigure}[b]{0.22\textwidth}
        \includegraphics[width=\textwidth]{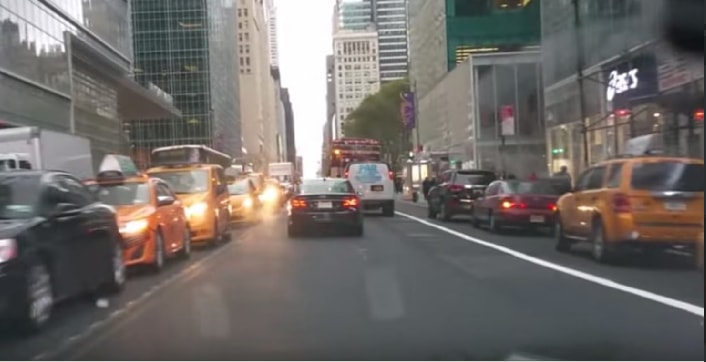}
        \caption{}
        \label{fig:Sequence_1}
    \end{subfigure}
    ~ 
    \begin{subfigure}[b]{0.22\textwidth}
        \includegraphics[width=\textwidth]{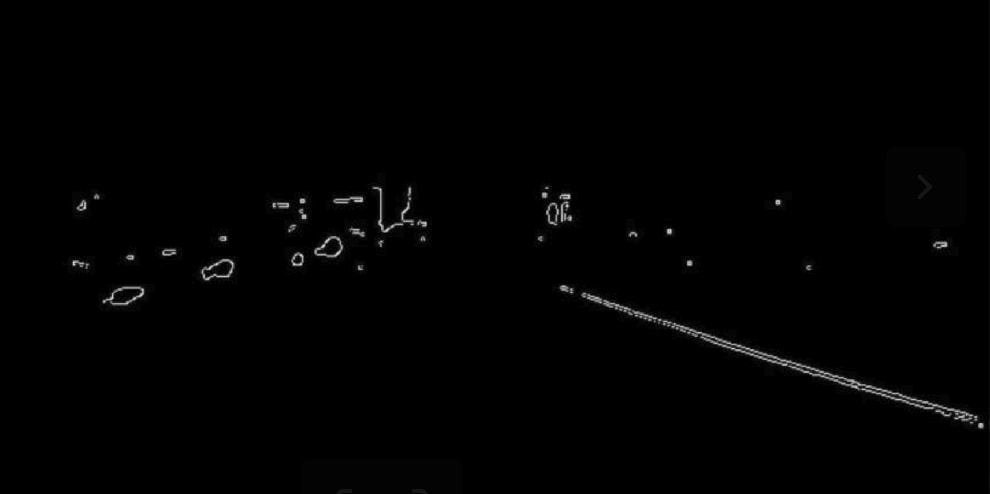}
        \caption{}
        \label{fig:Sequence_1_output}
    \end{subfigure}
    ~ 
    \begin{subfigure}[b]{0.22\textwidth}
        \includegraphics[width=\textwidth]{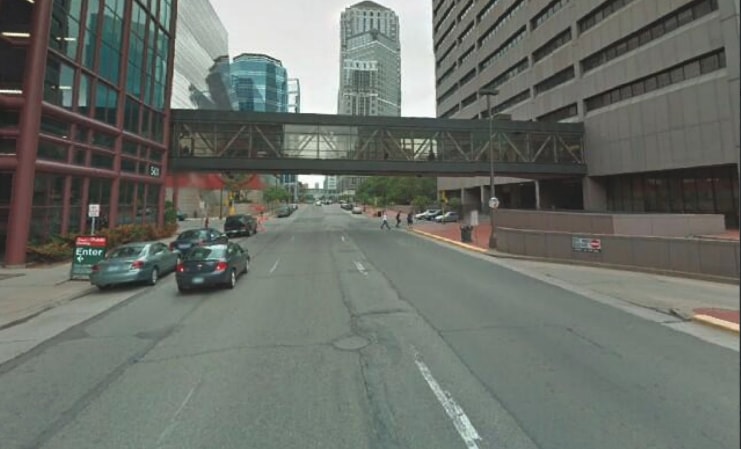}
        \caption{}
        \label{fig:Sequence_2}
    \end{subfigure}
    ~
    \begin{subfigure}[b]{0.22\textwidth}
        \includegraphics[width=\textwidth]{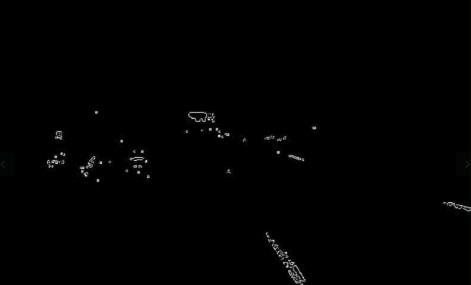}
        \caption{}
        \label{fig:Sequence_2_output}
    \end{subfigure}
    ~ 
    \begin{subfigure}[b]{0.22\textwidth}
        \includegraphics[width=\textwidth]{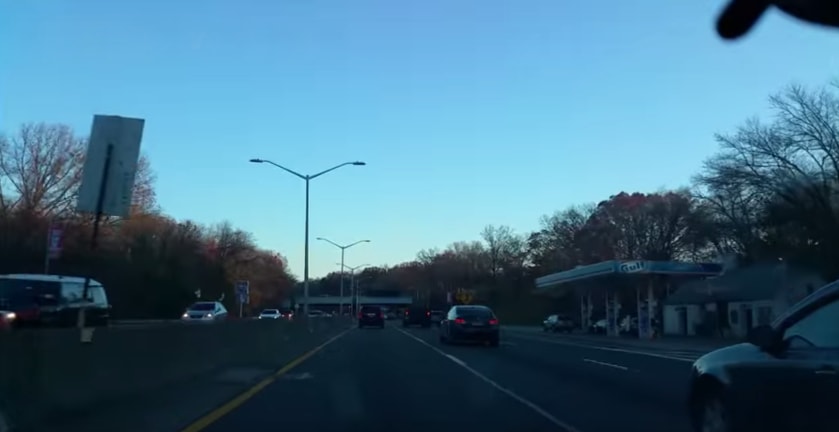}
        \caption{}
        \label{fig:Sequence_3}
    \end{subfigure}
    ~
	\begin{subfigure}[b]{0.22\textwidth}
        \includegraphics[width=\textwidth]{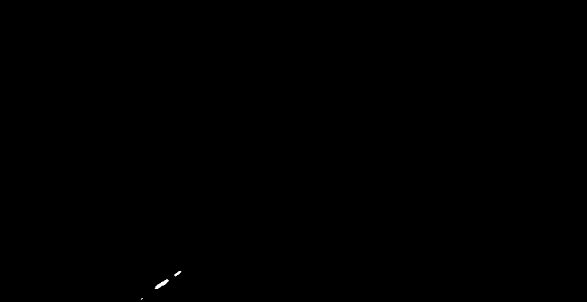}
        \caption{}
        \label{fig:Sequence_3_output}
    \end{subfigure}
    ~
    \begin{subfigure}[b]{0.22\textwidth}
        \includegraphics[width=\textwidth]{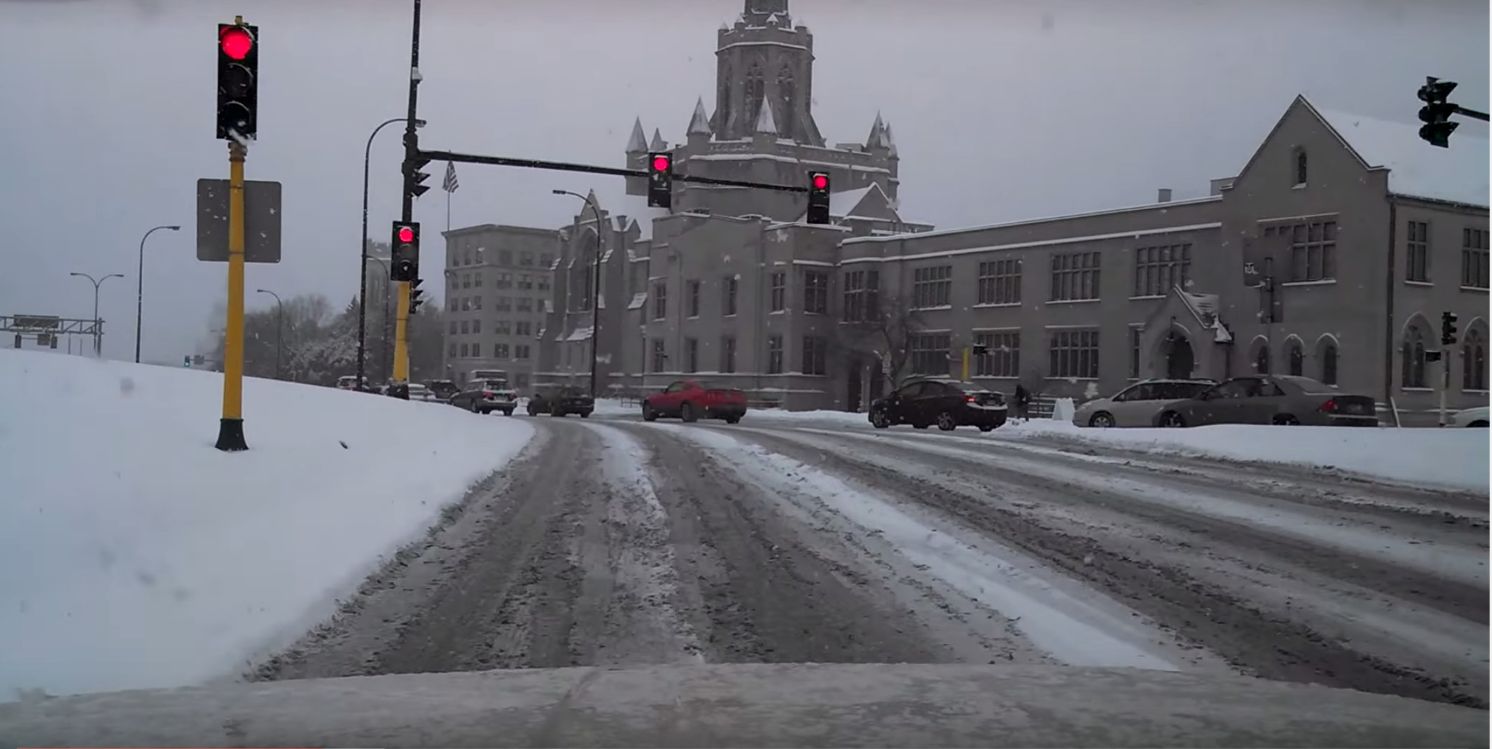}
        \caption{}
        \label{fig:Sequence_4}
    \end{subfigure}
    ~
    \begin{subfigure}[b]{0.22\textwidth}
        \includegraphics[width=\textwidth]{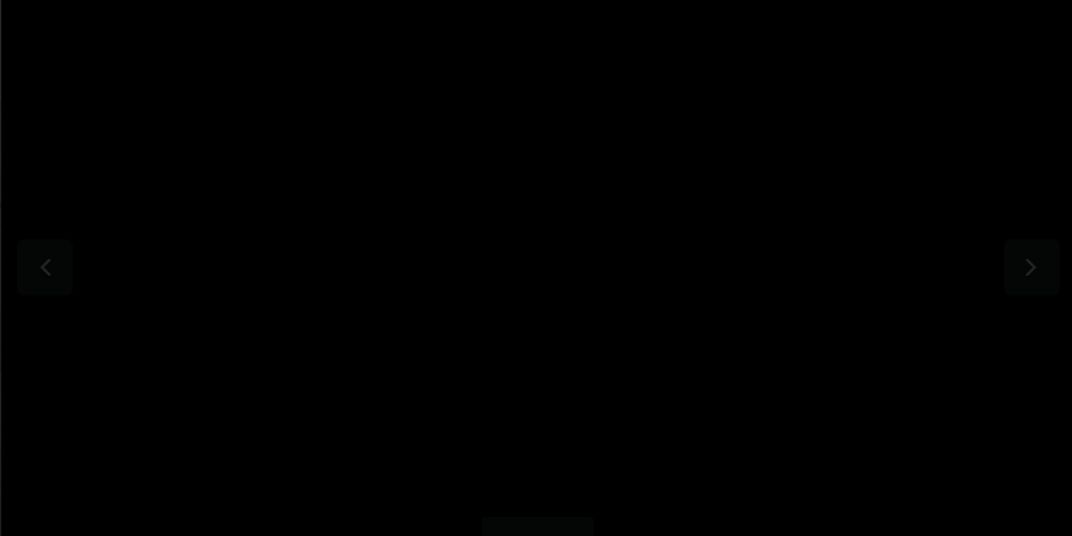}
        \caption{}
        \label{fig:Sequence_4_output}
    \end{subfigure}
    \caption{Visual lane tracking on several urban scenes from YouTube\texttrademark{} videos. Snapshot (\ref{fig:Sequence_1}) (output in (\ref{fig:Sequence_1_output})): lane markers not distinct in the center, though side markers are detectable. Snapshot (\ref{fig:Sequence_2}) (output in (\ref{fig:Sequence_2_output})): lane markers mostly washed out. Snapshot (\ref{fig:Sequence_3}) (output in (\ref{fig:Sequence_3_output})): evening drive, low-light conditions make the lane markers almost undetectable. Snapshot (\ref{fig:Sequence_4}) (output in (\ref{fig:Sequence_4_output})): snow-covered roads, no lane markers visible.}
\label{fig:Urban_Sequence}
\end{figure}

Across manufacturers (and in some cases, vehicle models), a variety of sensing methods are used in order to provide accurate detection of road features and consequently prevent traffic mishaps. Such sensors include but are not limited to laser scanners, radar, proximity sensors, and visible-spectrum cameras. Vision is an unobtrusive, low-cost, low-power sensor but requires appropriate lighting, unobstructed view, and fast processing time for deployment in autonomous and assisted driving applications. In spite of its disadvantages, vision sensors carry a strong appeal for deployment in mass production systems, mostly because of its low-power, inexpensive nature. For example, Subaru offers a stereo vision system termed EyeSight~\cite{SubaruEyesight} for lane detection and collision avoidance, with a stereo camera pair mounted on either side of the center rear-view mirror. This provides depth perception and lane tracking, and an intelligent drive system provides adaptive braking and cruise control, collision avoidance and lane-departure warnings. While the system has shown to work well in manufacturer testing, it is not immune to common failure cases, namely occlusion of road surfaces from snow, mud or a large vehicle, poor lighting condition, variable weather and ambiguity arising from feature confusion. An example scenario is show in Figure~\ref{fig:Urban_Sequence}, where a sequence of four snapshots demonstrate how changing conditions affect the quality of the center lane markers in the visual scene. The lane detection system is using a real-time segmentation approach; however, irrespective of the particular algorithm used to segment or detect the center lines, the input images themselves are of significantly degraded quality for robust lane tracking.

This paper proposes a system called SafeDrive\footnote{\url{https://github.com/jiawei-mo/SafeDrive}}, which is a significantly inexpensive approach for robust visual lane detection in severely degraded conditions, without relying on exotic, costly sensors which would be prohibitive for financial and compatibility reasons. Under poor visibility, the system uses the vehicle's location data to locate alternate image of the road ahead from an available ``road-view'' database. Once this image is acquired, a visual lane detection algorithm is applied to find the lane markers. Subsequently, visual feature matching is then applied to find corresponding image feature points between the ``current'' image and the alternate ``database'' image (which was acquired at some point in the past during significantly better visibility) to approximately find the location of the lane in the current image\nostarnote{JM: same as in abstract, image alignment(feature and pixel) is better}, even under zero lane visibility. For the development of this approach, an Android-based application called DriveData has been developed to capture a variety of data from a device mounted on (or even outside) the vehicle. The long-term goal is to provide an affordable solution to be used on a smartphone mounted on the windshield to provide lane departure warnings in extreme cases, provide safety recommendations under the current driving conditions based on visual, location and acceleration data.

\section{Related Work}
\label{sec:related}

A large body of literature exists on different aspects of autonomous driving and driver's assistance technologies, a number of which relies on robotics and computer vision methods. The Carnegie-Mellon NavLab~\cite{Thorpe_1985_2467} project has produced some of the earliest implementations of self-driving cars, and have extensively used vision for a number of subtasks, including road and lane detection~\cite{kluge1990explicit}. Stereo vision systems have been used for lane detection; \emph{e.g.}, in~\cite{nedevschi20043d,bertozzi1998gold}. Dedicated parallel processing for road and lane detection have been investigated in the GOLD~\cite{bertozzi1998gold} framework. Kluge et al.~\cite{kluge1995deformable} applied deformable shapes and Wang et al.~\cite{wang2004lane} used ``snakes'' for detection and tracking of lane markers. Spline-fitting methods for lane tracking have also been applied~\cite{wang1998lane}. Kim~\cite{kim2008robust} investigated robust lane tracking under challenging conditions with poor visibility and rapidly changing road traffic -- similar in nature to the problem we are addressing in this paper. However, that work does not consider the case of zero-visibility of lane markers, which we attempt to resolve. We use color-based segmentation and line-fitting in our work, and a number of authors have investigated similar approaches (\emph{e.g.},~\cite{chiu2005lane,gonzalez2000lane,borkar2009robust}). Often used with a combination Bayesian filtering and estimation methods, these methods have shown to work well under clear visibility conditions. Some researchers looked into the problem of rear-view lane detection~\cite{takahashi2002rear}. Interested readers are pointed to the paper by Hillel et al.~\cite{hillel2014recent} for an in-depth survey of recent advances in road and lane detection problems.
\section{Methodology}
\label{sec:methodology}
\begin{figure}[htp]
	\centering
    \includegraphics[width=0.6\linewidth]{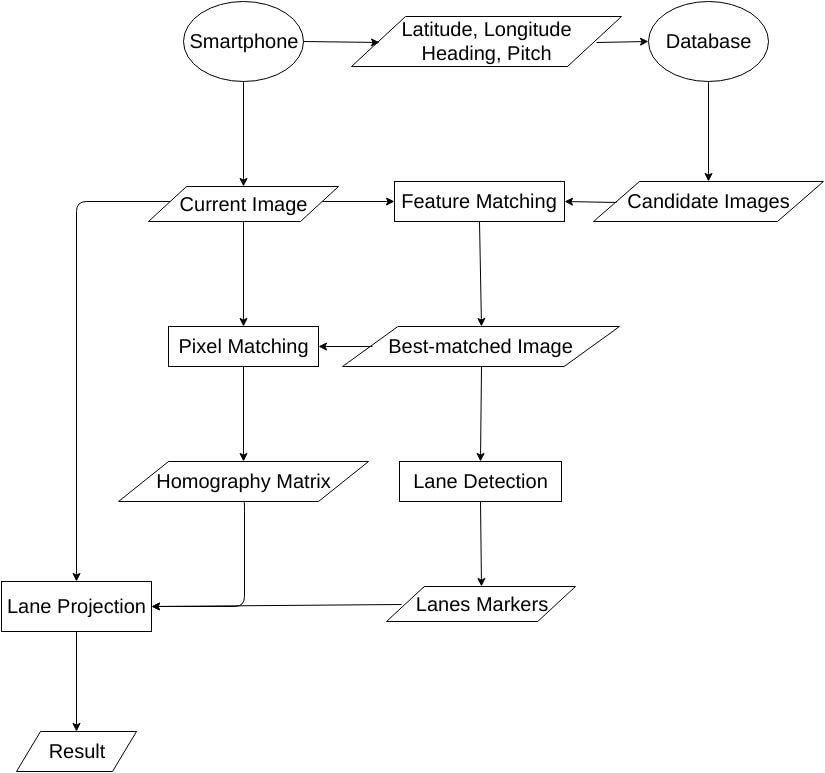}
    \caption{Diagram depicting various stages of the SafeDrive process.}
    \label{fig:flowchart}
\end{figure}

The core of our approach relies on robust alignment of the \emph{current} image (\emph{i.e.}, the image acquired from a smartphone device mounted on the windshield), where lane markers are not clearly visible, to an image in a database of road images taken at approximately the same location and heading, which has good visibility of lane markers. We rely on vehicle location data to retrieve alternate image of the correct location, with the proper heading and camera pitch angle\nostarnote{JM: camera angle(heading and pitch)?}, from a database of images. For the current implementation, we rely on imagery from the Google Street view data, and access it using the Google Maps API~\cite{GoogleStreetViewAPI}; however the database can be generated by the application itself during a previous traversal of the same path (the conjecture being people tend to travel a small subset of possible routes many times as part of their daily routine). Our method takes three sets of inputs: frames of video from a camera facing the road, orientations around the $x$, $y$, and $z$ axes, and a vehicle pose composed of latitude, longitude, and compass heading. The orientation data is used to acquire camera pitch angle. Images are indexed by [latitude,longitude] locations extracted from the Global Positioning System(GPS), and current vehicle heading\nostarnote{JS:L or bearing data. JM: bearing==pitch? bearing in robotics is more likely to be heading I think}. Depending on location, GPS data may be noisy (for reasons stemming from obstructions from tall buildings, or having no clear line-of-sight to satellites) and consequently it may not be possible to locate the proper image data corresponding to the actual vehicle location. We address this problem by finding a set of imagery data from the few closest locations to that being reported by the vehicle, and then using a feature-matching process to find the best match to the image being seen by the camera. This is possible as long as some road landmarks are visible, not the road or lane markers themselves. Once the image is found, it is then analyzed by a color-based segmentation algorithm and subsequently a line fitting process to find the lane. Position of the lane in the (poorly-visible) live frame is found through image alignment against the (clearly-visible) alternate imagery. The entire process is illustrated as Figure \ref{fig:flowchart}.

The image aligning process is done over two different steps to improve accuracy of lane marker detection and localization. The first step uses feature-based matching to choose the best image in database corresponding to the current image and vehicle location. Afterwards, a pixel-based matching approach enhances the alignment further by using the common regions found by the feature-based matching. The following sections describe the processes in detail.

\subsection{Feature-based Matching}
\label{sec:feature-match}

To correctly project the lane markers onto the current image, the proper image at the current location must be recovered from the database. Based on initial location data and camera angle, we need to find the best-matched image in the database to the current image, or equivalently, optimally matching the location and camera angle combination.
In the first step towards this goal, since latitude and longitude are highly correlated, a grid search approach is undertaken to find the closest latitude and longitude to the current vehicle location. Since we assume the initial guess for the camera angle is close to the true value, with latitude and longitude getting closer to the true value, the current image and candidate image from database will have more common content.

After finding the closest location, the next step is to find the closest camera angle. By \emph{camera angle}, we denote the \emph{heading} (\emph{i.e.} rotation around the vertical axis, or bearing) and \emph{pitch} (\emph{i.e.}, rotation about the axis perpendicular to the direction of the motion). Under the assumption that initial pitch guess is close to the true value, when pitch is fixed, the more accurate the heading is, the more overlapped content there will be. A similar argument applies for the pitch angle. Thus, the optimal heading angle and the optimal pitch angles can be searched separately. To find the best heading and pitch angles, a binary search strategy is adopted starting at an initial guess close to the actual angles, as recovered from the IMU of the smartphone. For example, with initial assumption of $150^{\circ}$, the program will search for the optimal angle (either pitch or heading) between $145^{\circ}$ and $155^{\circ}$ using binary search.

For our purposes, the criterion we use to measure ``similarity'' between two images is based on the number of matched feature points. This is because we are not after true similarity, as the current and candidate images may have very different appearances. The intent is to find two images with maximally overlapped visual content, which will essentially ensure the highest number of feature point matches for realistic scenes. Feature points are detected using the Harris corner detector\cite{harris1988combined}. Instead of keeping every point with large response, we force every point to be at least a certain distance away from any other point. This is to ensure an even spatial distribution of feature points to improve the likelihood of finding the most accurate image match. The ORB feature descriptor~\cite{rublee2011orb} is used to extract feature descriptors from both current image and candidate image, and matched between the current and candidate image using a \textit{K}-nearest-neighbors metric. For each feature point in the current image, two most similar points will be picked up from candidate image. If the matching distance of the most similar point is significantly smaller than that of the second-best point, (specifically, we set a minimum of 30 \emph{percent} reduction in distance), the most similar point is regarded as a potential match. Ideally, matching from current image to candidate image will have an identical result as the matching from candidate image to current image. However, because of visual distortions and occlusions, these two results are not always identical. To ensure an accurate match, descriptor matching is thus run again in the reverse direction (\emph{i.e.}, from the candidate image to the current image), in order to remove inconsistent matches. To further remove outliers , the RANSAC~\cite{fischler1981random} algorithm is applied to find the optimal homography between two images.

After running feature matching on all possible candidate images near initial guess, the one with the maximum matched points is chosen as the best-matched image. An approximate homography matrix is also obtained.

\subsection{Pixel-based Matching}
\label{sec:pixel-match}

The homography matrix obtained after running RANSAC on feature-based matches is not always accurate, especially when the feature matches are not spatially evenly distributed. For example, if most feature matches are located on the left half of the image, applying feature-based matching will project the markers on the left-half of the image accurately; however, the projections on the right-half will exhibit large discrepancies (see Figure~\ref{fig:feature_diff2} for an example). Depending on the scene, the spatial distribution of feature matches may be heavily skewed.

To further optimize the result, pixel-based matching is applied after feature matching. However, there are two challenges when applying pixel-based matching. The first challenge, as the best-matched image from the database was taken at a prior time, it will almost certainly contain different objects (\emph{e.g.}, cars, pedestrians, construction equipment). To accurately align the the best-matched image using pixel-based matching, such content must be ignored, and only common content shared by two images should be considered. Another challenge arises from photometric distortion. One object might have different pixel intensity values in images taken in different time(\emph{e.g.} higher intensity during a sunny day than a cloudy day). 

\begin{figure}
	\centering
    \includegraphics[width=\textwidth]{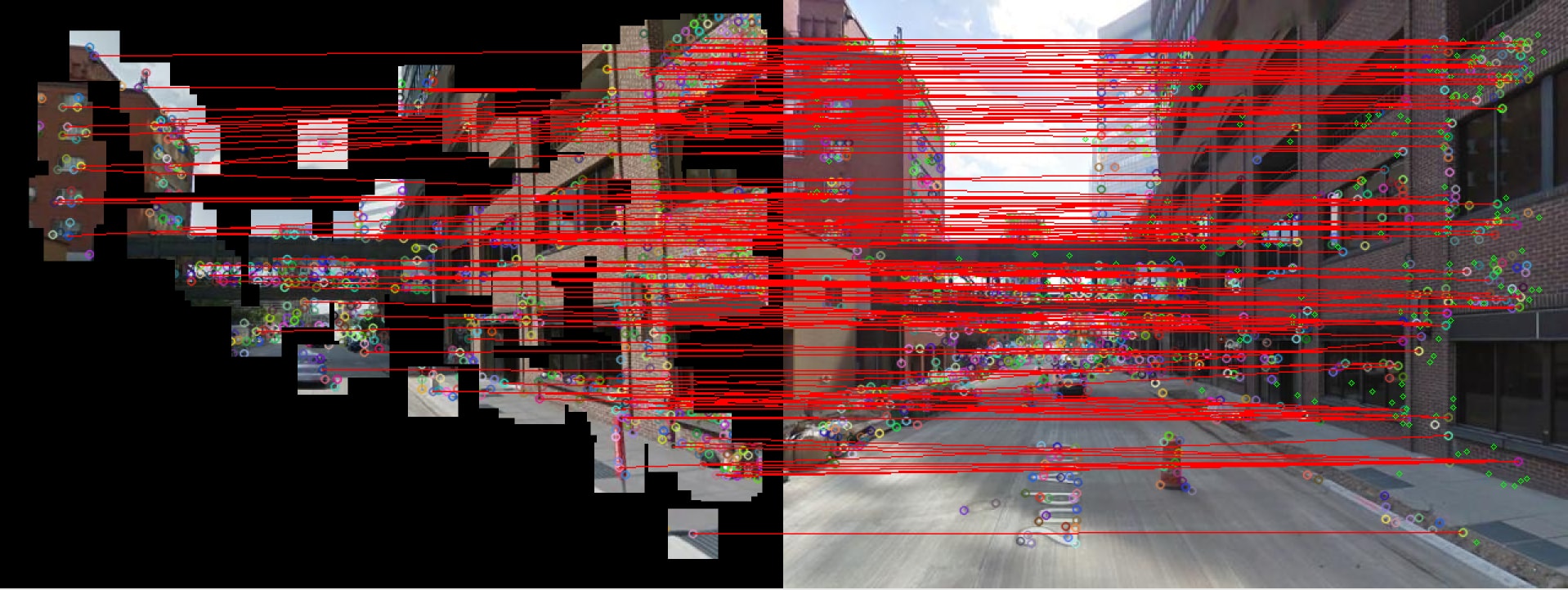}
    \caption{The process of extracting pixels with ``common'' visual content. The feature-based matching (in red lines) are used to choose the point features, and for each feature point, a square subwindow is extracted from the candidate image, centered on that feature point. Stitching together all these windows results in an image with most ``uncommon'' visual elements removed.}
    \label{fig:common_extraction}
\end{figure}

To extract common content between the best-matched image and current image, feature-based matching is first applied to find common points between two images, then a window centered at each matched point on the best-matched image is extracted(since the transformation from the best-matched image to current image is being sought). These windows are put together to form a filter, which filters out uncommon content between the images. One example is shown in Figure~\ref{fig:common_extraction}. Since the best-matched image has the maximum number of matched points among all candidate images in database, it will maximize the number of ``common'' pixels for pixel-based matching. 

The pixel-based matching algorithm we use is Enhanced Correlation Coefficient Maximization(ECC)\cite{evangelidis2008parametric}. Compared with other pixel-based matching methods, such as~\cite{szeliski2006image}, ECC achieves robustness against geometric and photometric distortions by normalizing pixel intensities. Besides, ECC is an area-based alignment which accepts a mask(window filter in our case) as input to specify a subset of pixels as region of interests. 

Finally, pixel-based matching outputs a refined homography matrix mapping from the best-matched image in the database to the current camera image. 

\subsection{Lane Marker Detection}
\label{sec:lane_detection}

Once the current image has been matched to a database image, the lane markers are detected on the best-matched image and overlaid on the current image. Lane markers are normally either white or yellow, pixels of other hues are first removed using a color-based segmentation algorithm. Any pixel whose RGB value ranges from (180,180,190) to (255, 255, 255), and from (0, 150, 170) to (150, 255, 255) are retained as white and yellow pixels respectively. As other objects with similar hues could be in the scene, a Canny line detector is used to find lines among those white/yellow pixels. Since most of lane markers have clear boundary, their outlines are detected and preserved, and most outliers are eliminated. These lane markers are then projected onto the current image according to the homography matrix calculated before. To make the final result appear realistic, instead of projecting lane markers outlines, all pixels in the vicinity of the lane marker outline are projected. A small number of outlier pixels may be projected onto the current image, but as real pixels are projected, the effect is negligible.


\begin{figure}[h!]
  \centering
  \begin{subfigure}[t]{0.3\linewidth}
    \centering
    \includegraphics[width=\linewidth]{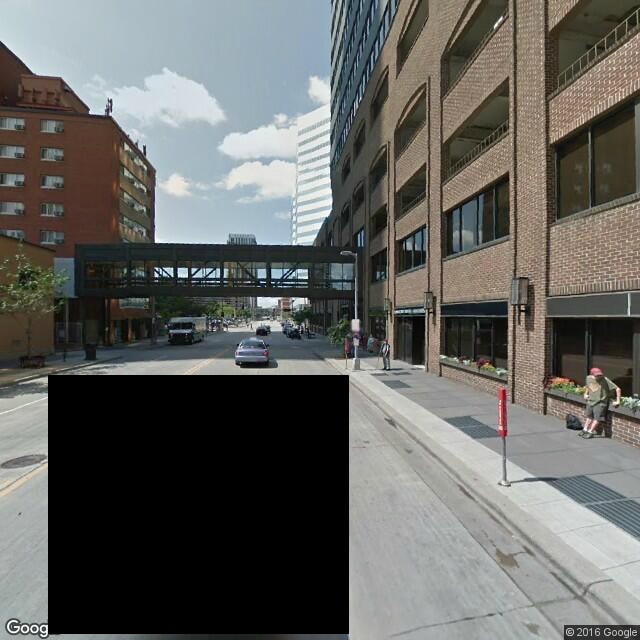}
    \caption{Current image taken from camera, artificially corrupted to remove lane markers.}
    \label{fig:current}
  \end{subfigure}
    ~ 
  \begin{subfigure}[t]{0.3\linewidth}
    \centering
    \includegraphics[width=\linewidth]{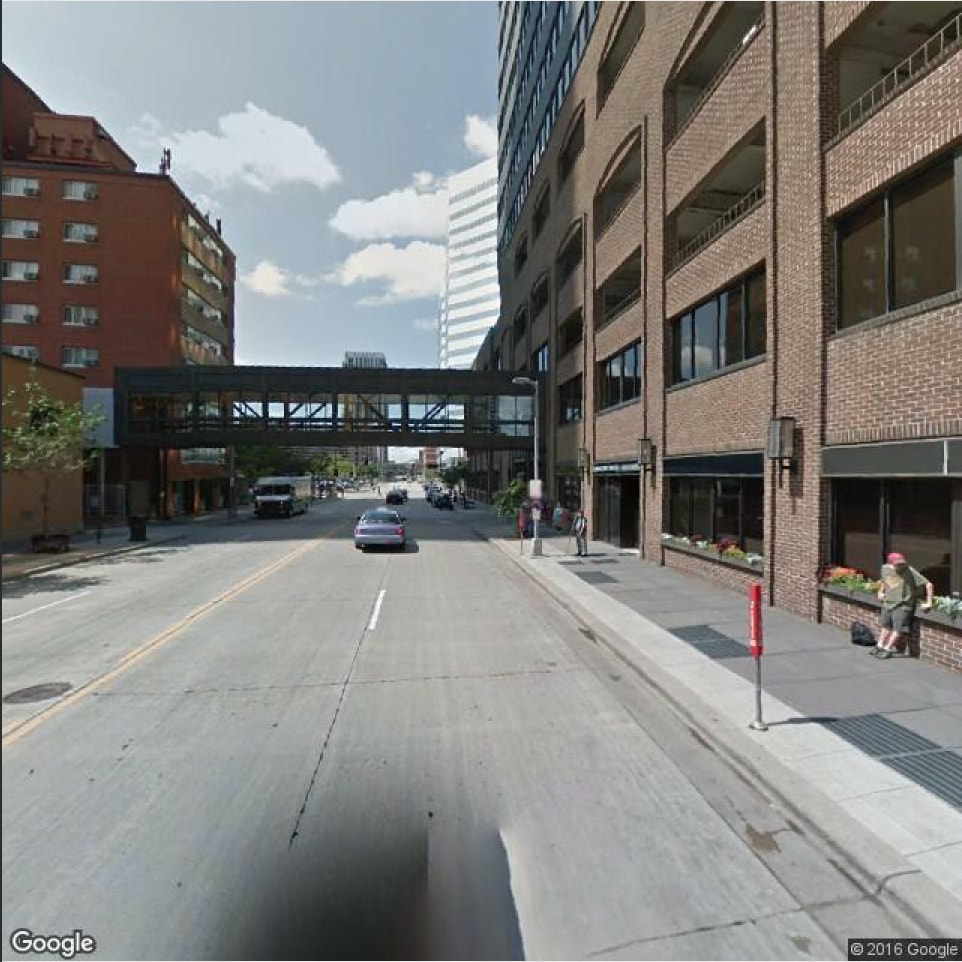}
    \caption{Best-matched image in database found by feature-based matching.}
    \label{fig:matched}
   \end{subfigure}
    ~ 
  \begin{subfigure}[t]{0.3\linewidth}
    \centering
    \includegraphics[width=\linewidth]{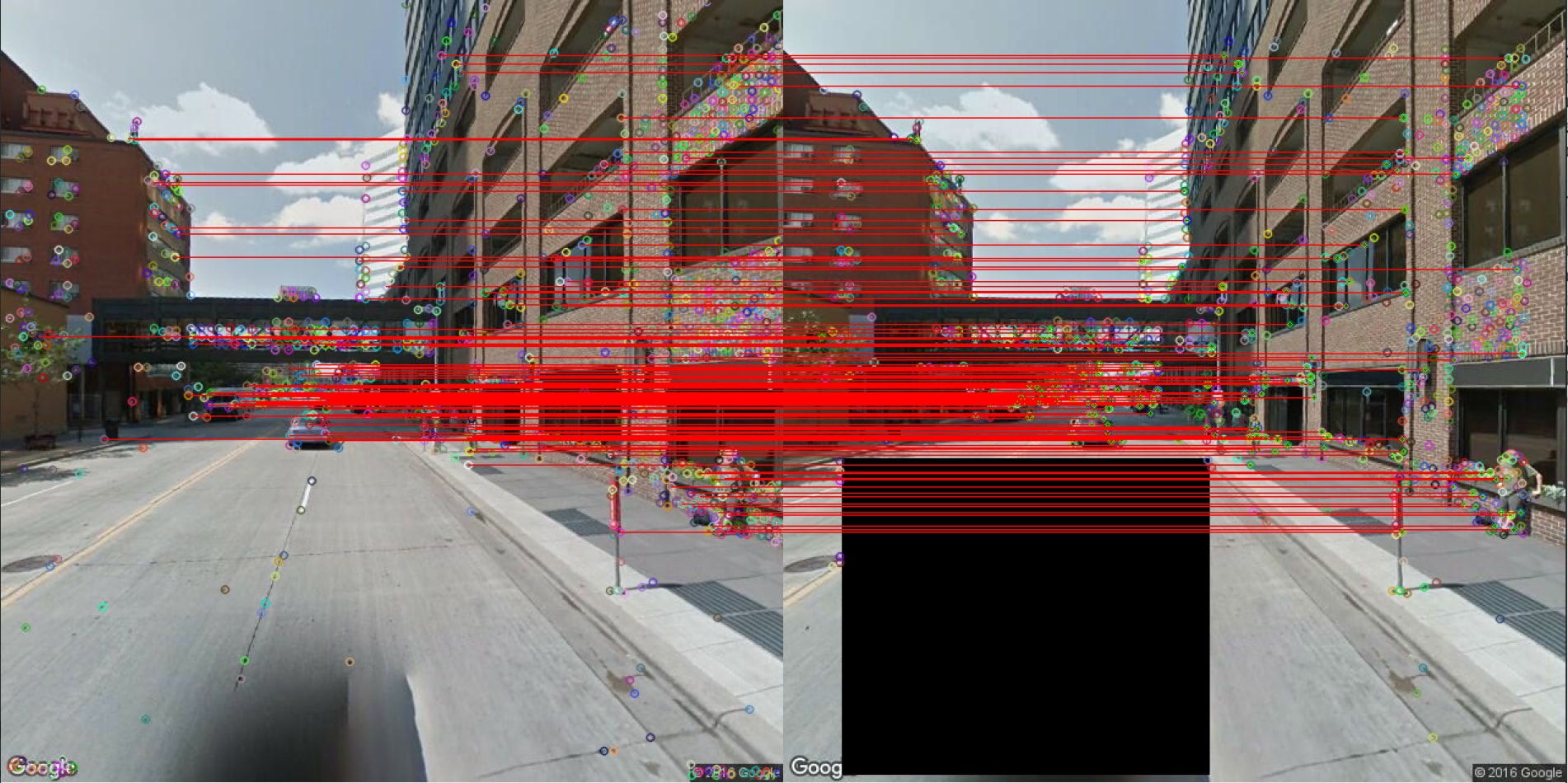}
    \caption{Feature matches between current and candidate image.}
    \label{fig:feature_match}
   \end{subfigure}
   ~
   \begin{subfigure}[t]{0.4\linewidth}
     \centering
     \includegraphics[width=\linewidth]{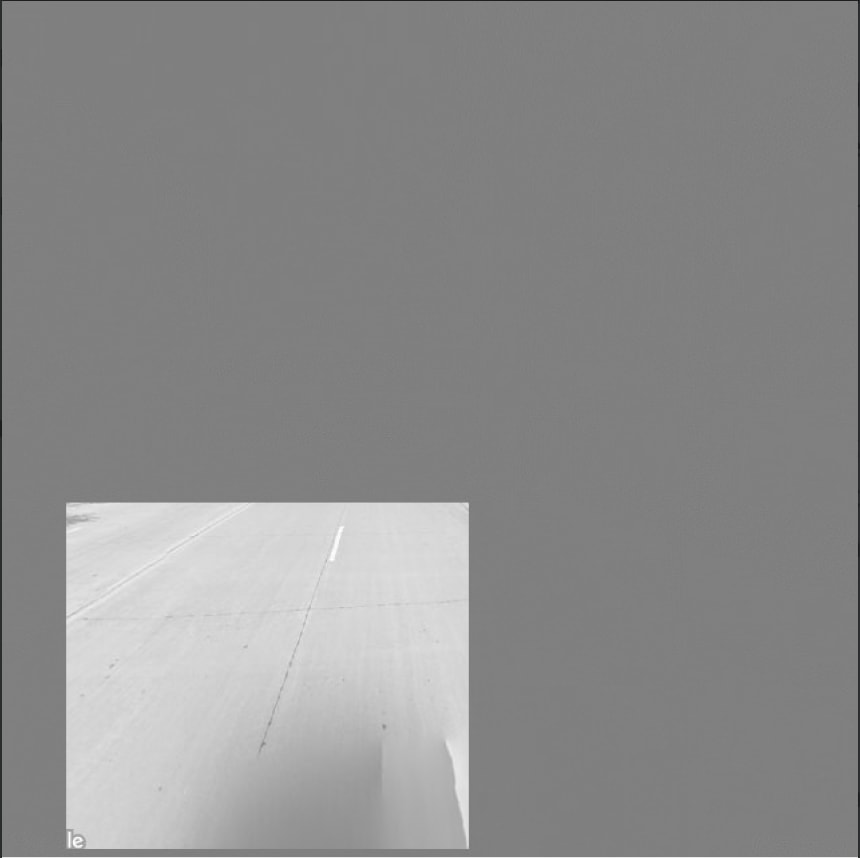}
     \caption{Pixel difference between projected best-matched image and current image. The gray sections indicate near-exact match between pixel intensities between the target and projected best-matched image.}
    \label{fig:feature_diff}
   \end{subfigure}
   \label{fig:simple_case}
     ~ 
  \begin{subfigure}[t]{0.4\linewidth}
    \centering
    \includegraphics[width=\linewidth]{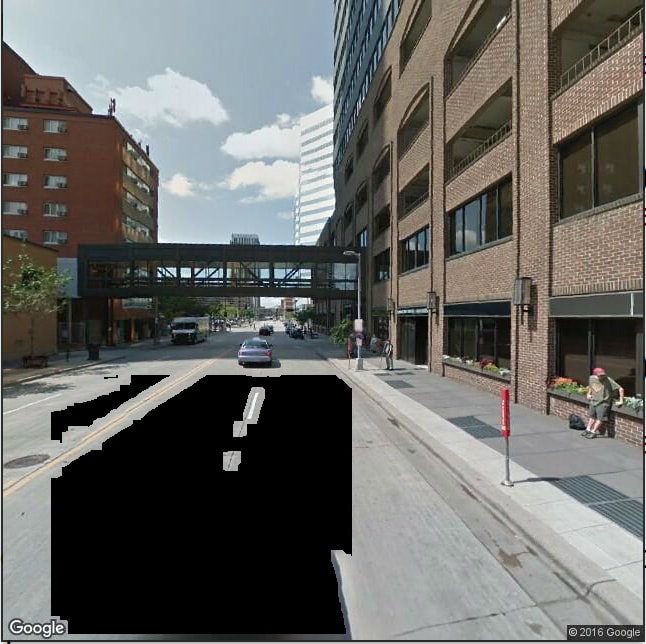}
    \caption{Projection of lane markers onto current image based on feature matching}
    \label{fig:feature_res}
   \end{subfigure}
   \caption{Experimental evaluations for the first case (see Section~\ref{sec:case_one}), where the view of road is artificially corrupted to simulate poor visibility. Pixel-based matching is not required in this case.}
\end{figure}

\section{Experimental Evaluation}
\label{sec:experiments}

An evaluation of SafeDrive's performance is presented below under two different scenarios. In the first, the view of road is artificially corrupted to simulate poor visibility of road markers and used as the current image. In the second, two different images of the same geographical location taken at two different times are used. 

\subsection{Evaluation $1$}
\label{sec:case_one}
\nostarnote{I put [H] below to fix figure position. Figures look bigger in downloaded PDF version, so I put them in one column.}

In the first case, the current image is downloaded from Google Street View, at the location of $(44.9745000^{\circ}, -93.2704977^{\circ})$ with camera heading of $218.36^{\circ}$ and pitch of $0^{\circ}$, taken in July 2015. As we are using the GSV of the same date as the database, the current image is guaranteed to exist in the database. To make road partially invisible, parts of road is painted black using photo editing software. The result is shown in Figure~\ref{fig:current}. Alongside with current image, we will also have an estimate of the initial location, including latitude and longitude, as well as an estimate of the initial camera angle  (heading and pitch). Using these initial estimates, we perform feature-based match, as described above. The feature matching process is illustrated in Figure~\ref{fig:feature_match}. After scanning through all possible images, the best-matched image is obtained with the maximum number of matched points, which can be seen in Figure~\ref{fig:matched}. In addition, the estimated current latitude/longitude/heading/pitch combination is $(44.9745000^{\circ}, -93.2704977^{\circ}, 218.36^{\circ}, 0^{\circ})$, which matches the true values exactly. Applying the homography matrix directly without performing a pixel-based match already produces quite accurate results, as can be seen in Figure~\ref{fig:feature_res}.

To illustrate the accuracy of the image alignment step, we compare the projected image from best-matched image with current image. Specifically, the target intensity of the output image is set at  $I_{output} = ((I_{projected}-I_{current}) / 2 + 128)$, where $I$ is the photometric intensity, to visualize the pixel difference. If the pixels are properly aligned, the output is effectively set to 50\% gray. In this test case, as can be seen in Figure~\ref{fig:feature_diff}, all pixels except for the artificially corrupted block appear gray, pointing to a near-perfect alignment. The sum of squared difference (SSD) of the intensity values across the entire image is $42603.4$, The corrupted part of the image is the main contributor for the high value of the SSD. The homography matrix is computed to be
\begin{equation}
  \begin{bmatrix}
      1.0000       & 1.5377e-15 		& -4.0880e-13 \\
      -1.1998e-16      & 1.0000 		& -1.9691e-13 \\
      -6.3838e-19  & -3.8465e-18 	& 1.000
  \end{bmatrix}
\end{equation}

which is almost identical to the identity matrix, further proving the accuracy of the result. 

\subsection{Evaluation $2$}
\label{sec:case_two}

The second test case illustrates a more realistic scenario, where the two images (current versus database image) would likely to have different content, albeit being taken at almost the exact location and with similar camera angles. To complicate matters further, it would be unlikely to find an image in the database with exactly the same location and camera angle. SafeDrive thus attempts to find the optimally closest values for location and camera angles. 

\begin{figure}[h!]
  \centering
   \begin{subfigure}[t]{0.23\textwidth}
     \centering
     \includegraphics[width=\textwidth]{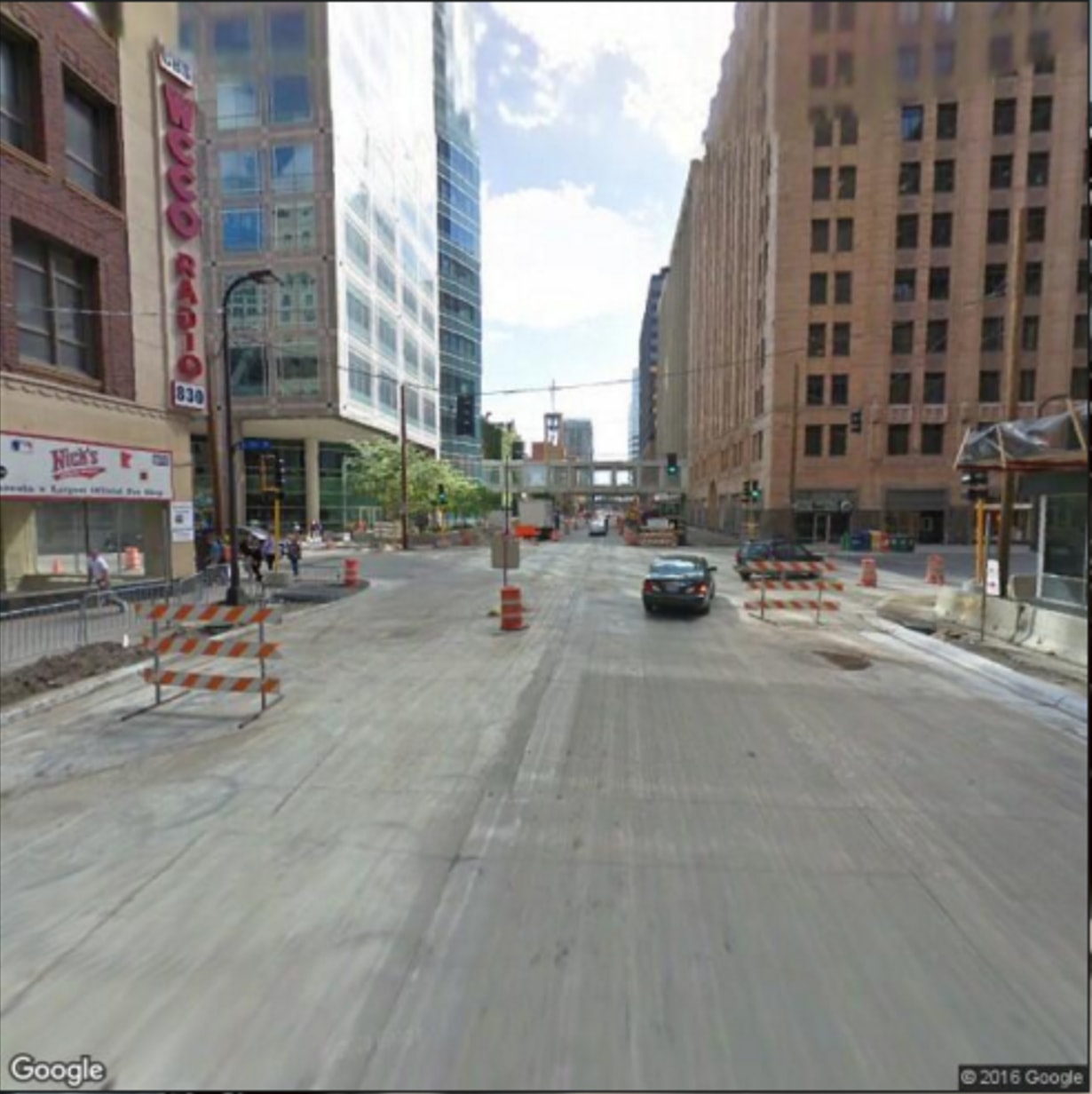}
     \caption{Current image taken from camera. Lane markers completely invisible.}
     \label{fig:current2}
   \end{subfigure}
   ~
   \begin{subfigure}[t]{0.23\textwidth}
     \centering
     \includegraphics[width=\textwidth]{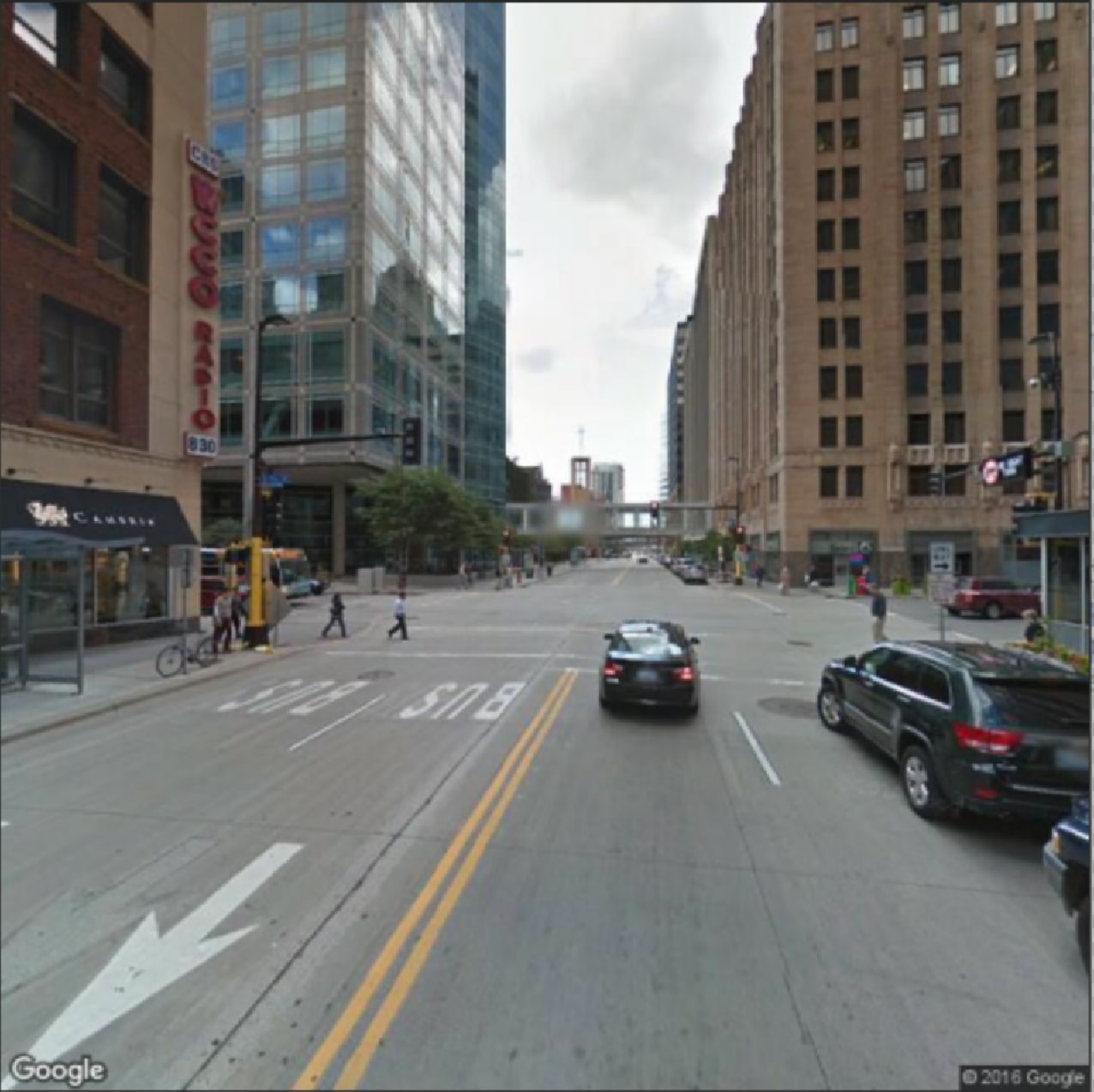}
     \caption{Best-matched image in the database found by feature-based matching. Note that the current image is \emph{not} in the database.}
     \label{fig:matched2}
   \end{subfigure}
   ~
  \begin{subfigure}[t]{0.23\textwidth}
    \centering
    \includegraphics[width=\textwidth]{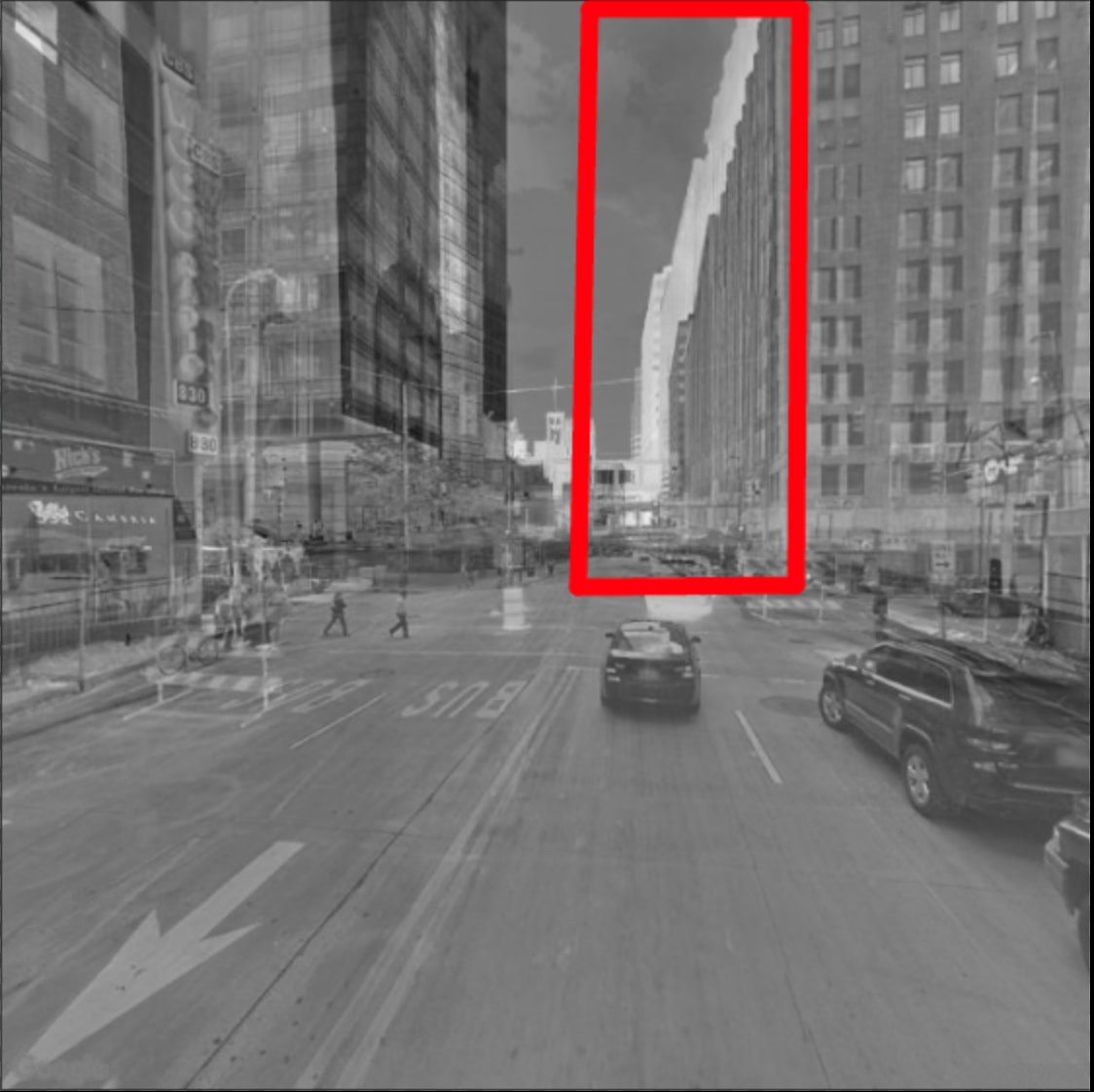}
    \caption{Large projection errors (\emph{e.g.}, region marked with the red rectangle) after feature matching.}
    \label{fig:feature_diff2}
   \end{subfigure}
    ~
   \begin{subfigure}[t]{0.23\textwidth}
     \centering
     \includegraphics[width=\textwidth]{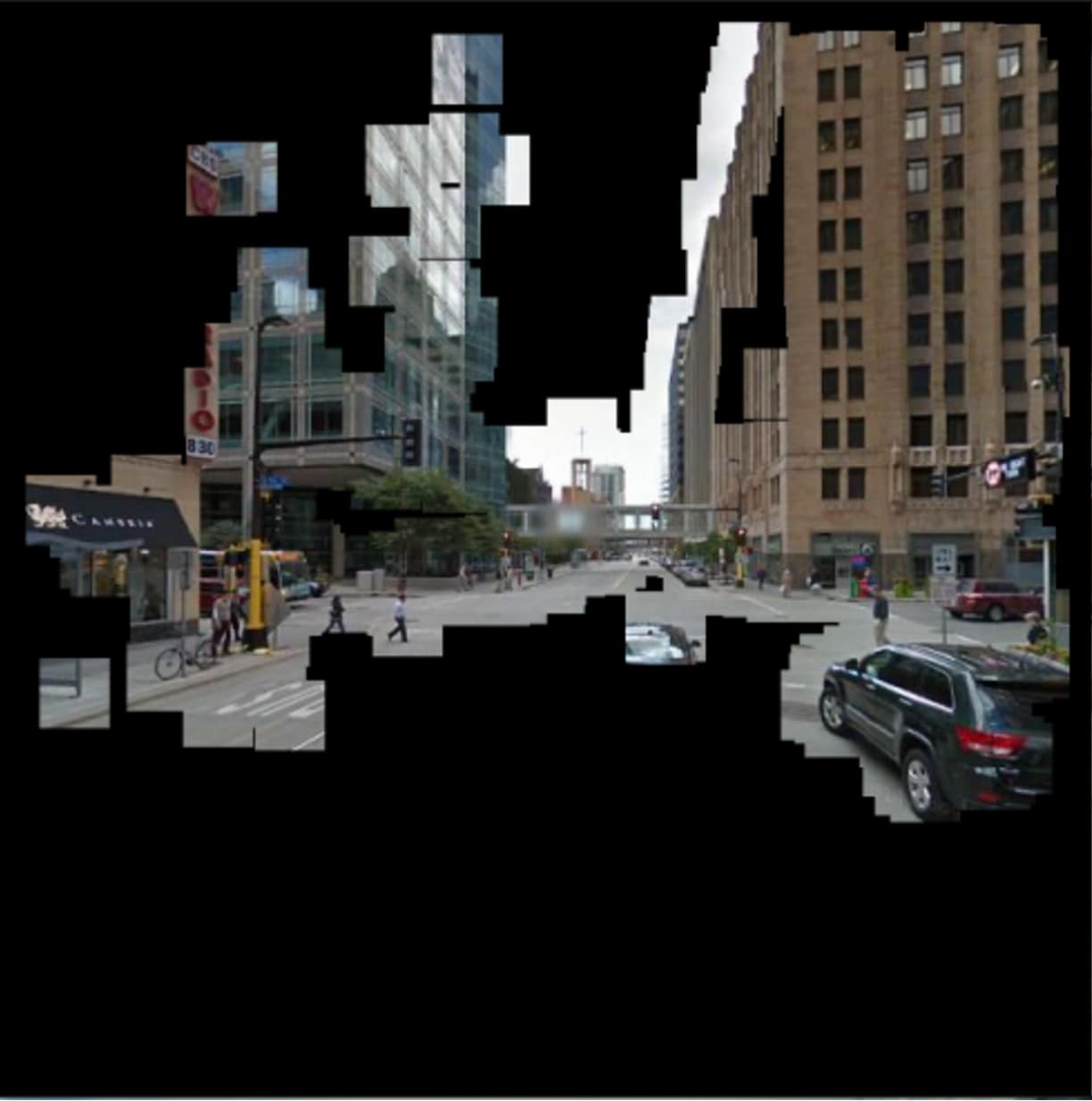}
     \caption{Output image after extracting common pixels guided by common feature point matches.}
     \label{fig:filtered_img}
   \end{subfigure}
   ~
   \begin{subfigure}[t]{0.23\textwidth}
     \centering
     \includegraphics[width=\textwidth]{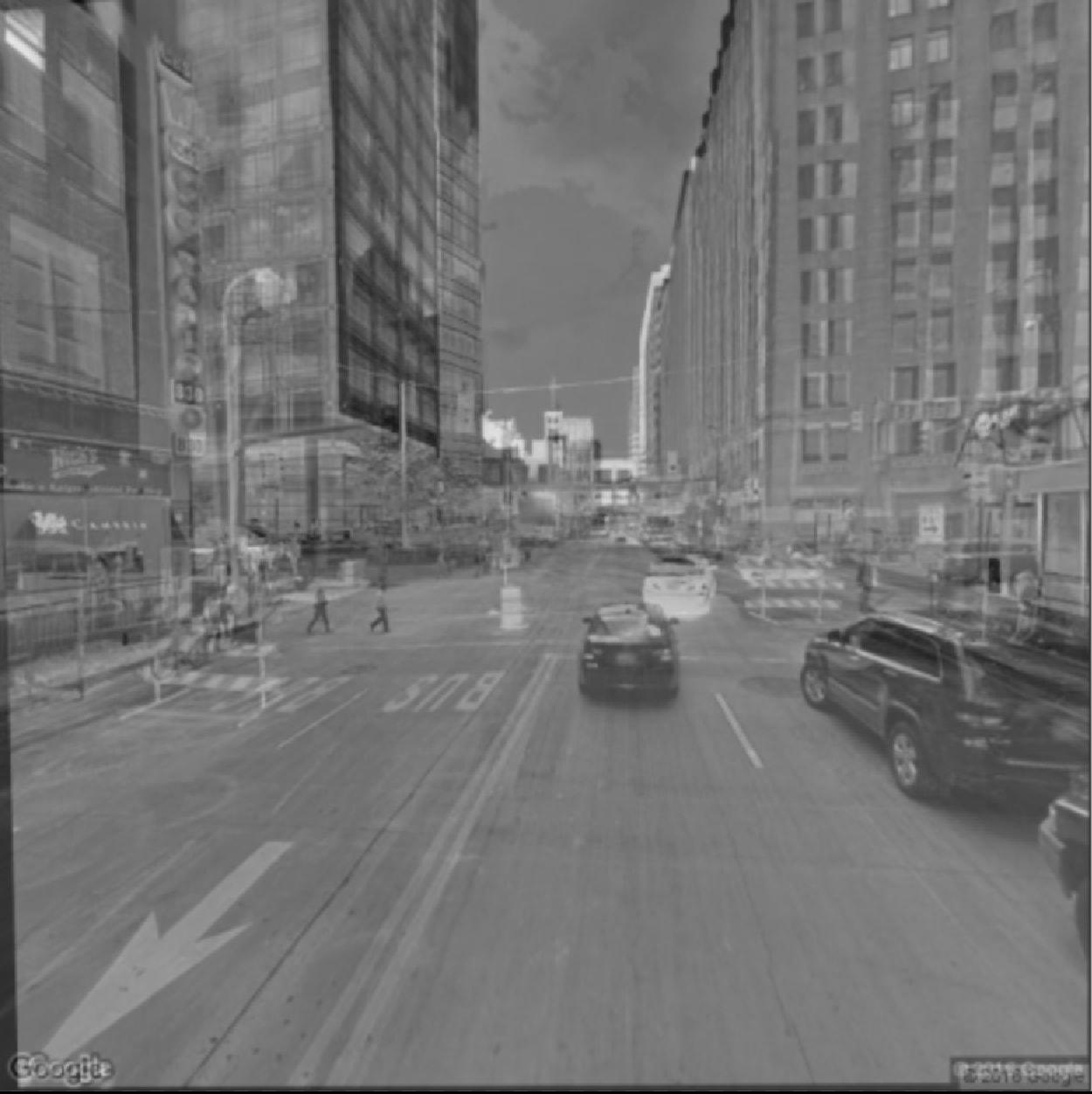}
     \caption{Pixel difference between projected best-matched image and current image after pixel-based matching}
     \label{fig:final_diff}
   \end{subfigure}
    ~
   \begin{subfigure}[t]{0.23\textwidth}
     \centering
     \includegraphics[width=\textwidth]{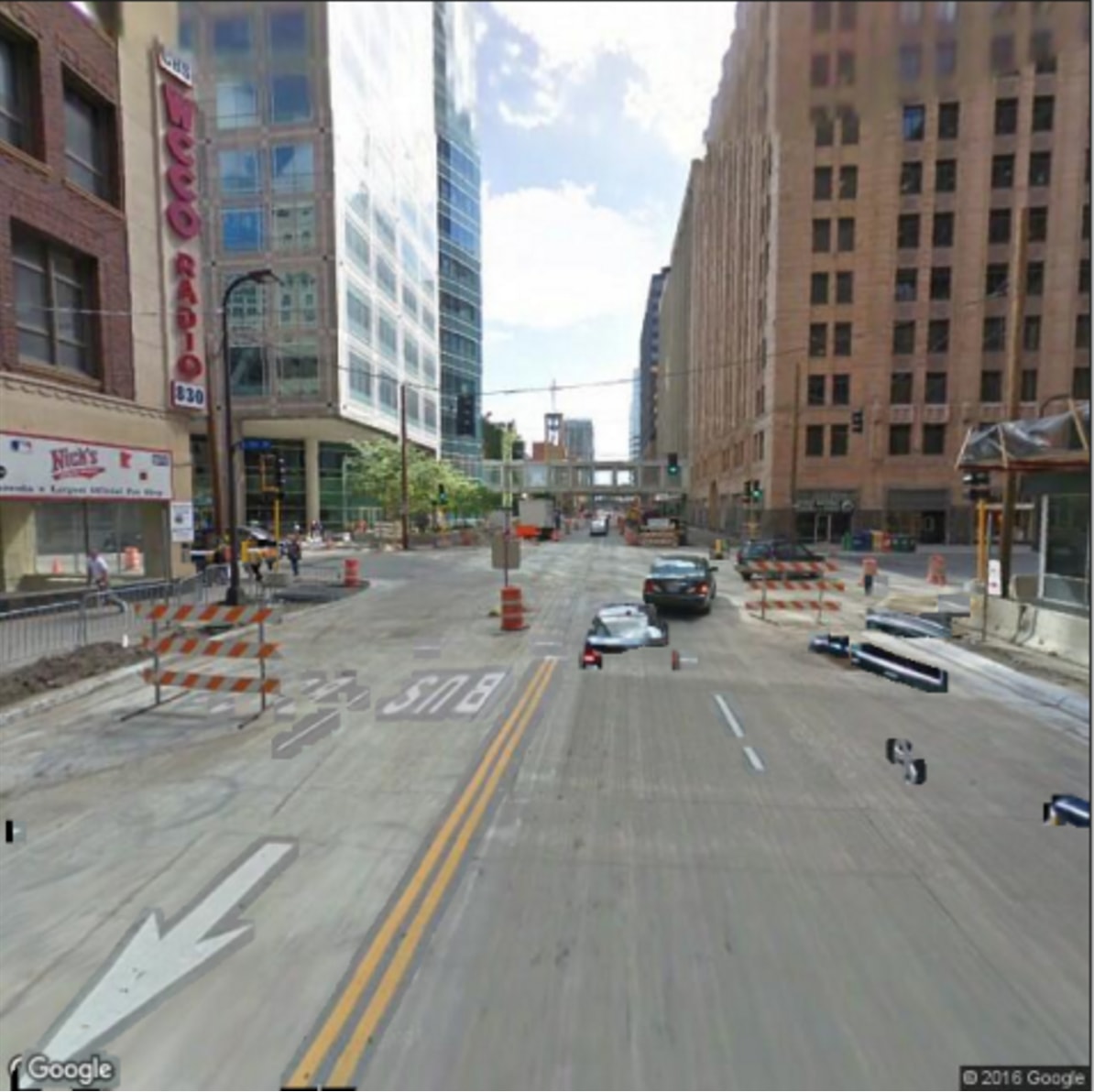}
     \caption{Final result after projecting lane markers onto current image by pixel-based matching.}
     \label{fig:final_res}
   \end{subfigure}
   \label{fig:complex_case}
   \caption{Experimental evaluations for the second case (see Section~\ref{sec:case_two}), with two different images of the same geographical location taken at two different times. There are significant differences in spite of both images looking at the same scene.}
\end{figure}

In this case, we force the current image to be not included in the search database. Specifically, parameters of the current image, which is shown in Figure~\ref{fig:current2}, are  (latitude/longitude/heading/pitch) = $(44.9759631^{\circ}, -93.269327^{\circ}, 195.74^{\circ}, 0^{\circ})$, and has been taken in July 2009. While the database images were taken in the September of 2014. 
The best-matched image in the database is found with parameters $(44.9759^{\circ}, -93.2694^{\circ}, 200.74^{\circ}, 0^{\circ})$, which is shown in Figure~\ref{fig:matched2}. Please note the database image is different from the current image but essentially contains the view of the same location taking at an approximately similar heading. Figure~\ref{fig:feature_diff2} is the pixel difference when we run feature-based matching only. The SSD value of the pixel intensity values between the current and projected database image is 33387.4. In comparison, the SSD value for the first case discussed in the previous section is $42603.4$. In that case, the corrupted part of the image is the main contributor for the high value of the SSD. A lower SSD value for a realistic scenario is a good indicator for the effectiveness of our approach. The homography matrix is:

\begin{equation}
  \begin{bmatrix}
      0.9234       & -0.0817 		& 17.9786 \\
      -0.0064      & 0.9536 		& -24.0934 \\
      -8.8504e-05  & -7.5111e-05 	& 1.000
  \end{bmatrix}
  \label{eq:fea_homo2}
\end{equation}

As can be seen in Figure~\ref{fig:feature_diff2}, there is significant drift between current image and best-matched image, which can be noticed by the building borders marked in red rectangle. To minimize drift, pixel-based matching is applied to refine the homography matrix obtained from feature-based match. Before applying pixel-based matching, content not shared by both images are eliminated. We extract a $41\times 41$ window centered at each matched feature point on the best-matched image. The image after applying the feature-based window filters is shown in Figure~\ref{fig:filtered_img}. Even though some outliers remain (\emph{e.g.}, the car on bottom right), most content on the filtered image are shared by both images. The pixel-based matcher is then executed on the filtered image. The final pixel difference after pixel-based matching is shown in Figure~\ref{fig:final_diff}. The SSD value is further lowered to $31576.6$.

The refined homography matrix is:
\begin{equation}
  \begin{bmatrix}
      0.9171        & 0.0239       & -3.510 \\
      -0.0947       & 1.0134       & 7.705 \\
      -0.0001   	& 4.237e-05    & 1.000
  \end{bmatrix}
  \label{eq:final_homo}
\end{equation}
\\
which is is different from the previous matrix (\emph{i.e.}, matrix~\ref{eq:fea_homo2}), signifying the changes imparted by the pixel-based matcher. As the best-matched frame was taken at a different location with different camera angle, even though it is the closest one, neither matrix~\ref{eq:fea_homo2} nor matrix~\ref{eq:final_homo} is thus identity matrices. Compared to Figure~\ref{fig:feature_diff2}, the building borders in Figure~\ref{fig:final_diff} are better aligned as well.

Figure~\ref{fig:final_res} is the final result after projection. All lane markers, including double yellow lane, white dashed lane, bus-only sign, and direction arrow, are correctly detected and projected.

\subsection{Performance}
\label{sec:performance}
SafeDrive was developed in C++ and tested on a PC running Ubuntu $16.04$ with Intel Core i7 6700HQ CPU. OpenMP\cite{openmp08} has been used to run grid search concurrently to find the optimal latitude/longitude, but otherwise the code has not been optimized for performance. The search grid size is $3\times 3$. We run $5$ iterations searching for heading and 5 iterations searching for pitch. The maximum iteration of the ECC pixel matching algorithm is set to $50$. On average, the total process takes approximately $8$ seconds.


\section*{Acknowledgment}
The authors are grateful to Michael Fulton for his assistance, particularly in developing the DriveData\footnote{\url{https://github.com/fultonms/drivedata}} Android\texttrademark application and subsequent collection of a large volume of driving data. 

\section{Conclusions}
We have presented\nostarnote{JM: should we use 'presented'?} an algorithm for visual lane detection and tracking under poor visibility conditions, and even in cases the road surface is barely visible. This approach leverages the availability of alternate imagery of the same location and the ability to perform lane tracking in such imagery, eventually mapping the lane detection back to the original camera image. With sufficiently robust visual lane-finding algorithms, accurate pose detection, and robust methods to relate the past image with the live frame, we believe this algorithm can significantly improve driver safety. The ultimate goal for our work is to create an affordable system, and simultaneously improve the quality of autonomous transportation and occupant safety in road-going vehicles. Ongoing research is focusing on improved feature matching for lane location correspondence, compressed data handling and optimization for enhanced performance, and extensive testing on  data collected from a diverse set of geographic locations. 



\bibliographystyle{plain}
\small{
\bibliography{citation,allbibs}
}
\end{document}